\documentclass[runningheads]{llncs}
\usepackage{xcolor}
\usepackage{booktabs}
\usepackage{siunitx}
\usepackage{booktabs} 
\usepackage{threeparttable} 
\usepackage{multirow,multicol}
\usepackage{float}
\usepackage{graphicx}
\usepackage{siunitx}
\usepackage{amsmath}
\usepackage{fontawesome5}
\begin{document}
\title{Unleashing the Power of Pre-trained Encoders for Universal Adversarial Attack Detection}


\author{Yinghe Zhang\inst{1} \and
Chi Liu \inst{1}\faIcon{envelope} \and
Shuai Zhou \inst{1} \and
Sheng Shen \inst{2} \and
Peng Gui \inst{3}
}
\authorrunning{Y. Zhang et al.}
%
\institute{Faculty of Data Science, City University of Macau, Macao SAR, China \and
Design and Creative Technology Vertical, Torrens University Australia, NSW, Australia \and
School of Computer Science and Engineering, Wuhan Institute of Technology, Wuhan, China\\
\faIcon{envelope} Corresponding author: \email{chiliu@cityu.edu.mo}
}
%
\maketitle              
%
\begin{abstract}
Adversarial attacks pose a critical security threat to real-world AI systems by injecting human-imperceptible perturbations into benign samples to induce misclassification in deep learning models. While existing detection methods, such as Bayesian uncertainty estimation and activation pattern analysis, have achieved progress through feature engineering, their reliance on handcrafted feature design and prior knowledge of attack patterns limits generalization capabilities and incurs high engineering costs. To address these limitations, this paper proposes a lightweight adversarial detection framework based on the large-scale pre-trained vision-language model CLIP. Departing from conventional adversarial feature characterization paradigms, we innovatively adopt an anomaly detection perspective. By jointly fine-tuning CLIP's dual visual-text encoders with trainable adapter networks and learnable prompts, we construct a compact representation space tailored for natural images. Notably, our detection architecture achieves substantial improvements in generalization capability across both known and unknown attack patterns compared to traditional methods, while significantly reducing training overhead. This study provides a novel technical pathway for establishing a parameter-efficient and attack-agnostic defense paradigm, markedly enhancing the robustness of vision systems against evolving adversarial threats.

\end{abstract}

\section{Introduction}
Adversarial attacks, which inject human-imperceptible adversarial perturbations into a benign sample to fool a deep learning model, have emerged as a major type of attack in artificial intelligence. Extensive research has demonstrated that deep learning models, ranging from conventional small ones \cite{zhou2022adversarial} to emerging large multi-modality models \cite{yang2024new}, exhibit significant vulnerabilities against adversarial examples. Adversarial attack has become a considerable security threat in reality. For examples, attackers can exploit adversarial images to mislead autopilots \cite{eykholt2018robust}, or evade face recognition systems \cite{sharif2016accessorize}. 

The imperative need to mitigate adversarial attacks has spurred interest in defensive techniques. Adversarial example detection offers a promising first line of defense by scanning adversarial examples at a very early stage \cite{zhou2022adversarial}. Most of the recent detection methods rely on sophisticated frameworks and elaborate feature engineering. For example, Feinman et al. \cite{1-6} employ Bayesian uncertainty estimation to distinguish anomalous samples. Metzen et al. \cite{1-4} identify differences between clean images and adversarial examples in activation patterns. Ma et al. \cite{1-8} leverage Local Intrinsic Dimensionality for detection. Despite these methods having demonstrated certain defensibility in controlled environments, challenges remain that limit their broader applications. Detectors heavily relying on specific features are prone to overfitting, resulting in poor out-of-distribution performance against unknown or stealthier attacks \cite{1-5}. Moreover, developing these sophisticated detectors requires substantial prior knowledge of adversarial features, as well as significant cost, engineering expertise, and a large, diverse set of adversarial examples. 

\begin{figure}
    \centering
    \includegraphics[width=\linewidth]{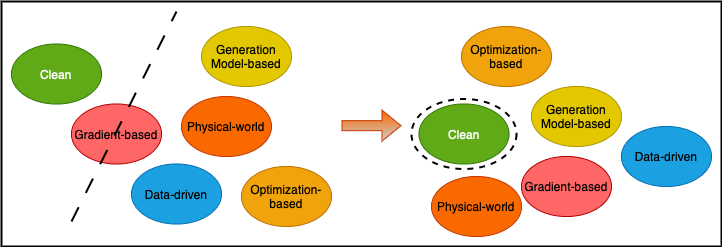}
    \caption{Static decision boundary isolation vs dynamic distribution generalization}
    \label{fig:motivation}
\end{figure}

To overcome the current limitations, more generalizable and lightweight solutions are needed. Unlike previous methods that explore detectable features of adversarial examples, we derive our motivation from an abnormity detection perspective, which aims to characterize the compact distribution of clean natural images. As shown in Fig \ref{fig:motivation}, previous methods learn a fixed decision boundary that separates benign images from known adversarial types but struggle with unseen ones. In contrast, modeling the distribution boundary of clean images naturally enables to distinguish between benign images and diverse adversarial examples, including unknown ones. 

However, directly modeling the distribution of natural images via supervised learning is often infeasible, primarily due to the need for massive and diverse datasets of natural images. To circumvent this limitation, we leverage the capabilities of emerging large pretrained encoders, such as the CLIP encoder \cite{3-1-1}, which is pretrained on millions of high-quality natural images. This pretraining endows CLIP with the capacity to generate compact, foundational representations of natural images. Consequently, we propose a methodology for transferring the pre-existing knowledge embedded within pretrained CLIP encoders to the task of adversarial example detection through efficient fine-tuning. Our approach incorporates a fusion model that concurrently fine-tunes both the text and vision encoders of a pretrained CLIP model, employing a trainable adapter network and prompt tuning, respectively. Evaluations conducted across eight common types of adversarial examples demonstrate that, even when trained on a limited number of instances of a specific adversarial type, our method achieves remarkably high detectability on previously unseen adversarial examples. This is accomplished with approximately $5\%$ of the parameters and $25\%$ of the training epochs compared to those required by a standard ResNet18 model.

The main contributions of this study can be summarized as follows:
\begin{itemize}
    \item We develop a dual-stream CLIP architecture combining prompt tuning and adapter modules, achieving superior adversarial detection accuracy with dramatically reduced parameter overhead compared to conventional CNN-based detectors.

    \item Our curriculum learning framework demonstrates enhanced generalization capability across both known and unseen attack patterns, significantly outperforming feature-engineering approaches in robustness metrics.

    \item We establish a comprehensive adversarial face detection benchmark incorporating diverse gradient-based attacks with multi-level perturbations, enabling holistic evaluation of vision-language models' defensive capabilities.
\end{itemize}

\section{RELATED WORKS}
\subsection{Adversarial Attacks and Adversarial Example Detection}
Adversarial attacks and adversarial example detection constitute two mutually constraining core directions in the field of deep learning security\cite{zhou2022adversarial}. On one hand, adversarial attack techniques attempt to bypass model decision boundaries by generating specifically perturbed samples, such as gradient sign-based perturbations in FGSM \cite{A1}, iterative optimization-based perturbations in PGD \cite{A3}, and generative perturbation-based methods in AdvGAN \cite{2-2-6}. On the other hand, adversarial example detection techniques, including dimensionality reduction and compression in Feature Squeezing \cite{2-3-1}, reconstruction error analysis in MagNet \cite{A5}, and confidence calibration in ODIN \cite{2-3-4}, aim to identify such anomalous samples through input feature analysis or model response monitoring.

For instance, when addressing highly transferable black-box attacks like MI-FGSM \cite{A5} and DIM \cite{2-2-5}, detection methods require integration of cross-model feature alignment (Zeroth-Order Optimization (ZOO) framework \cite{2-3-5}) or dynamic perturbation suppression (Local Intrinsic Dimensionality (LID) analysis \cite{1-8}). When confronting physical-world attacks such as adversarial patches in AdvFaces \cite{2-2-7}, detection technologies must incorporate multi-scale material consistency verification (Physical Attack Detection \cite{2-3-6}).

\subsection{Pre-trained Models and Fine-Tuning Methods}
Pre-trained vision-language models (CLIP \cite{3-1-1},  PLDG \cite{yan2024prompt}) achieve zero-shot transfer capability through large-scale cross-modal alignment, enabling them to perform diverse visual tasks ( ImageNet classification \cite{3-1-1}) via text prompts without fine-tuning, while demonstrating cross-domain robustness. For scenarios requiring domain-specific adaptation, fine-tuning pre-trained models (Devlin et al., BERT \cite{2-4-1}, Radford et al. \cite{2-4-2}) has emerged as the dominant paradigm for enhancing task performance by transferring knowledge from large-scale pre-training to downstream tasks. However, traditional fine-tuning faces challenges such as high computational costs and catastrophic forgetting, driving the development of parameter-efficient methods: Adapter Modules (Houlsby et al.)\cite{2-4-4} enable efficient adaptation by inserting lightweight task-specific layers, preserving the general representation capabilities of pre-trained models while significantly reducing resource requirements. This technical framework provides a unified optimization paradigm for domain adaptation in both vision-language models and text-only models.

\section{METHODOLOGY}
Adversarial example detection serves as a critical task for ensuring the robustness of deep learning models. The core objective of detection systems is to determine whether an input image is (a) a clean image --- untouched by adversarial perturbations --- or (b) an image contaminated by adversarial attacks. Traditional detection methods face three primary challenges: (1) weak generalization performance; (2) separation of feature extraction pipelines, which struggles to capture multimodal correlations; and (3) computational efficiency bottlenecks in real-time detection scenarios. This section introduces our methodology based on the CLIP (Contrastive Language-Image Pre-training) model, which enables three detection strategies through progressive transfer learning.


\subsection{CLIP’s Generalization Advantage and Transfer Learning}
The CLIP model constructs a highly aligned visual-semantic feature space through contrastive learning on massive multimodal data (400 million image-text pairs) \cite{3-1-1}. This pre-training paradigm endows CLIP with unique generalization capabilities: its visual encoder captures globally relevant context (e.g., object structure, texture patterns) associated with semantics, while its text encoder distills abstract cross-modal concepts (e.g., "natural noise" vs. "adversarial perturbations") under natural language supervision. CLIP’s pre-trained features exhibit robustness against unseen attack types (e.g., diffusion-based perturbations \cite{3-1-2} or physical-world adversarial samples \cite{3-1-3}) by avoiding overfitting to specific local patterns (e.g., FGSM’s linear perturbations \cite{3-1-3}) and relying on semantic consistency for authenticity judgment \cite{3-2-1-8}.

This study aims to leverage CLIP’s generalization potential while optimizing its performance for adversarial facial image detection. Through systematic ablation experiments, we propose three progressive transfer learning strategies: Adapter, Prompt Tuning, and a fusion architecture.

\begin{figure}[!htbp]
    \centering
    \includegraphics[width=0.89\textwidth]{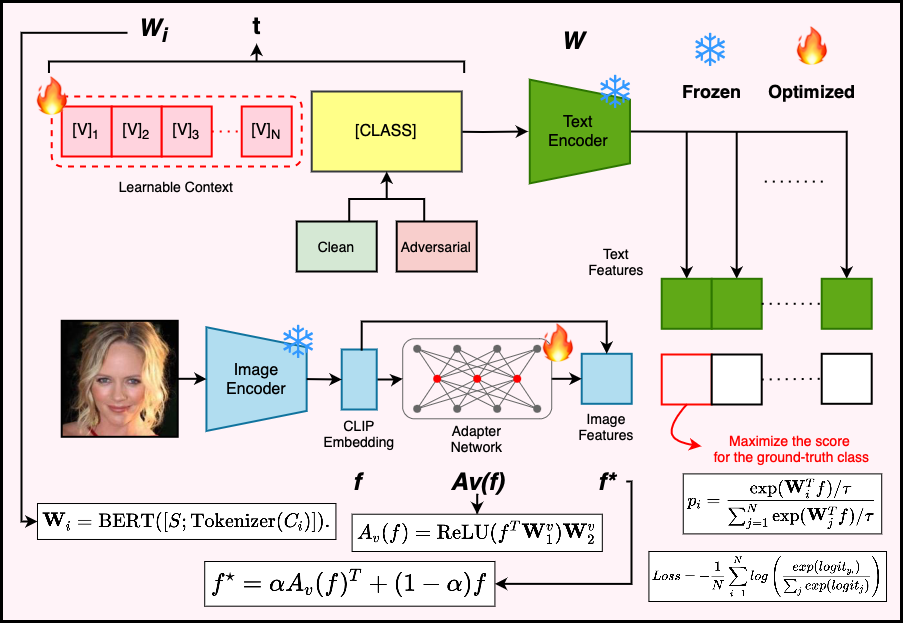}
    \caption{Flowcharts of different fine-tuning methods for pretrained Clip encoders: i) Visual Adapter; ii) Prompt Tuning; iii) our fusion architecture that incorporates Visual Adapter and Prompt Tuning for lightweight and generalized adversarial example detection, as shown in the black dash box.
    Visual adaptor or prompt tuning can be viewed as a simplified version of our fusion model, where only one modality is fine-tuned while the other remains frozen.
    }
    \label{flowchart}
\end{figure}

\subsubsection{3.2.1 Visual Adapter}
(Fig. \ref{flowchart})
As a visual enhancement component of the CLIP-Fusion model, the CLIP-Adapter introduces a lightweight adapter network at the end of the visual encoder \cite{3-2-1-8}. This adapter dynamically adjusts feature distributions via residual connections, with a core mechanism targeting high-frequency noise characteristics of adversarial perturbations \cite{3-2-1-9}. Specifically, it learns sparse convolution kernels to amplify channels correlated with abnormal frequency-domain responses (e.g., high-frequency components). Notably, Adapter achieves parameter efficiency --- requiring only fine-tuning a small fraction of parameters compared to full-model fine-tuning --- while preserving CLIP’s generic representational capacity and enhancing sensitivity to local perturbations \cite{3-2-1-8}.

In the current task, we only introduce and fine-tune a feature adapter in the image branch while keeping the text branch frozen. Specifically, we first utilize the unmodified CLIP backbone network to extract the input image's feature vector $f$  , ensuring the preservation of initial effectiveness and stability in image feature extraction.

Subsequently, through the fine-tuned image feature adapter $A_v(\cdot)$, (Equation\eqref{eq4})we perform transformation and fusion operations on the image features obtained$f$. The image feature adapter$A_v(\cdot)$ consists of a two-layer linear transformation with residual connections that integrate the transformed features with the original features, (Equation\eqref{eq5})ultimately yielding the enhanced image features $f^{*}$. This process introduces a small number of learnable parameters to enhance the model's adaptability to specific tasks while preserving the integrity of pre-trained knowledge in the original CLIP model.

\begin{equation}
    A_v(f)=\mathrm{ReLU}(f^TW_1^v)W_2^v
    \label{eq4}
\end{equation}

\begin{equation}
    f^*=\alpha A_v(f)^T+(1-\alpha)f
    \label{eq5}
\end{equation}

Finally, fixed text features $\mathbf{W}_{i}$ (Equation\eqref{eq6}) and newly generated image features $f^{*}$
are jointly used to calculate prediction probabilities in different categories. The text features $\mathbf{W}_{i}$ are generated by CLIP's text encoder and remain frozen throughout the entire process, thereby ensuring that the knowledge embedded in the text branch remains undisturbed.

\begin{equation}
    W_i=\mathrm{BERT}(\mathrm{Tokenizer}([H;C_i]))
    \label{eq6}
\end{equation}

\subsubsection{3.2.2 Prompt Tuning}
As the semantic guidance component of the CLIP-Fusion model, this strategy draws inspiration from CoOp(Context Optimization) \cite{3-1-2} by optimizing learnable textual prompts (Prompts) to activate abnormal semantic responses in CLIP’s text encoder. Its main function is to optimize the prompts, better align them with the objectives and enable the model to adapt to specific downstream tasks.(Fig. \ref{flowchart}) Its mechanisms include:

In our task, we adopt the transfer learning strategy of Context Optimization (CoOp) from Zhou et al.  \cite{3-1-2}, then fine-tune the model for adversarial face image detection. First, we achieve dynamic semantic alignment by designing learnable prompt templates (e.g., ``This image is corrupted by $\{$perturbation type$\}$'') and map adversarial visual features to the abnormal semantic space via contrastive learning. Second, to realize cross-attack generalization, prompts adaptively adjust during training, covering multiple attack patterns without explicitly defining perturbation types (e.g., PGD  \cite{A3})  \cite{3-1-2}.
In CoOp, learnable vectors are combined with contextual words in prompts. These vectors are initialized either randomly or with pre-trained word embeddings. During training, only these vectors are optimized, while the text and visual encoders remain frozen.

\begin{equation}
    t=[V]_1[V]_2[V]_3\ldots[V]_N[CLASS]
    \label{eq:1}
\end{equation}

In Equation~\eqref{eq:1}, each $[V]_n$ ($n \in \{1, \ldots, N\}$) is a vector with the same dimension as word embeddings, e.g., 768 in CLIP (ViT Large) \cite{3-1-1}. $N$, a hyperparameter indicating the number of contextual tokens (i.e., $[V]_1$, $[V]_2$, $[V]_3$, ..., $[V]_N$), was set to $[4, 8, 16, 24]$ in our experiments. $[CLASS]$ represents dataset class labels, such as "Clean" and "Adversarial". In each prompt $t_i$, the class label is replaced with the corresponding class name's word embedding. The prompt $t$ is input into the text encoder and optimized during training using the cross-entropy loss (Equation \ref{eq:2}).

\begin{equation}
    Loss=-\frac{1}{N}\sum_{i=1}^Nlog\left(\frac{exp(logit_{y_i})}{\sum_jexp(logit_j)}\right)
    \label{eq:2}
\end{equation}

\subsubsection{3.2.3 Fusion Architecture}
The visual adapter and semantic prompt tuning are then integrated using a multi-model dual-path architecture, as shown in Fig. \ref{flowchart}. Particularly, two strategies are used: 1) \textbf{Visual-Semantic Interaction: } Local perturbation features (e.g., gradient anomalies \cite{2-3-1}) extracted by the visual adapter and abnormal semantic descriptions (e.g., "potential adversarial noise") generated by prompt tuning interact through a cross-modal attention layer, producing joint discriminative features. 2) \textbf{Decision-Level Fusion: }The system combines confidence scores from both paths (adapter output and prompt-tuned output) using weighted averaging to suppress false positives from individual pathways. Experiments (Chapter \ref{chapter:experiment}) demonstrate significant accuracy improvements in the fusion strategy compared to individual approaches.

\section{Experiment}
\label{chapter:experiment}

\subsubsection{General Setup}
To systematically assess detection capability, we employ Accuracy and Macro-F1 scores as evaluation metrics, with classification thresholds calibrated on a 10\% validation subset. The training framework leverages the ViT-L/14 backbone pretrained on 9,999 samples, processing $224\!\times\!224$ RGB images through 16-sample batches. We implement stochastic gradient descent with momentum $0.9$, initial learning rate $2\!\times\!10^{-3}$, and weight decay $5\!\times\!10^{-4}$ , adopting 1-epoch linear warmup followed by cosine learning rate scheduling over 8 training epochs.
\\
The data augmentation pipeline incorporates random resized crops([0.08,1.0]), horizontal flips , and Gaussian blur compression(GB\_K:21), with pixel-level normalization using CLIP statistics(PIXEL\_MEAN: [0.481,0.458,0.408], PIXEL\_STD: [0.269,0.261,0.276]). All models are initialized with deterministic weights (SEED:17) and employ mixed-precision training (fp16), logging metrics every 5 iterations. Experiments are conducted using PyTorch 2.4.1 with CUDA 11.5 acceleration on NVIDIA RTX 4090 GPUs.

\subsubsection{Attack Methods and Datasets}
\begin{figure}[htbp]
    \centering
    \includegraphics[width=1\textwidth]{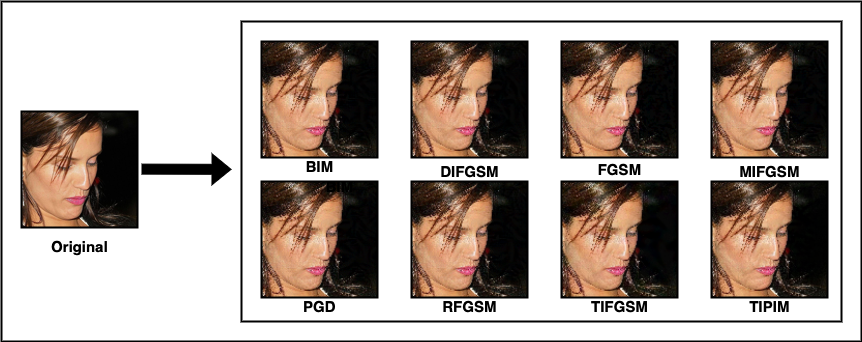}
    \caption{Visual comparison between natural images and eight adversarial examples. The perturbations are human-imperceptible.}
    \label{fig:perturbed_samples}
\end{figure}

We utilize eight recent gradient-based adversarial attack techniques: FGSM \cite{A1}, BIM \cite{A2}, PGD \cite{A3}, RFGSM \cite{A4}, MIM \cite{A5}, DIM \cite{A6}, TIM \cite{A7}, and TIPIM \cite{A8}. The gradient-based attacks are carried out on the CelebA-HQ datasets to produce adversarial faces. Fig. \ref{fig:perturbed_samples} shows examples for visual comparison. The perturbations are imperceptible, indicating the significant challenge of adversarial example detection.  

Our training dataset consists of two parts: $1999$ benign natural images sourced from the CelebA dataset, with an image resolution of $256 \times 256$, and 8,000 adversarial images generated using the above adversarial attack methods, with each contributing 1,000 images. 


\subsubsection{Baselines}
To systematically validate the effectiveness of the multimodal fusion framework, this study designed two baseline approaches based on ablation experiments: (1) Single-modality visual pathway (retaining only the Adapter module) for enhancing visual representation; (2) Single-modality semantic pathway (adopting only the dynamic prompt tuning mechanism) to optimize the semantic space. Through comparative ablation analysis, the experiments revealed the limitations of the visual-only pathway in fine-grained semantic perception and the bottlenecks of the semantic-only pathway in cross-modal alignment. These findings highlight the advantages of the proposed progressive fusion framework in achieving cross-modal collaboration and dynamic complementarity.

\subsection{Detectability}
As shown in the Table \ref{tab:combined_results}, the detection performance of the Adapter, Prompt Tuning, and Fusion Model architectures under various adversarial attacks is compared. All three architectures were evaluated using the same training set (9,999 images) and testing set (106,344 images). The experimental results indicate that the Fusion Model consistently achieves the highest accuracy and macro-F1 scores across most attack types. For example, under BIM, FGSM, PGD, RFGSM, DIFGSM, MIFGSM, and TIFGSM attacks, the Fusion Model attains an accuracy of approximately 99.87\% and a macro-F1 score of about 99.51\%. Under TIPIM attacks, the Fusion Model reaches an accuracy of 99.795\% and a macro-F1 score of 99.217\%, outperforming both the Adapter (with TIPIM values of 99.767\% and 99.107\%, respectively) and Prompt Tuning (with TIPIM values of 99.435\% and 97.789\%, respectively). These findings demonstrate that the dual-path architecture of the Fusion Model, which integrates both visual and semantic features, exhibits more stable and superior robustness against adversarial attacks, thereby significantly enhancing the generalization capability of adversarial attack detection.



\begin{table}[htbp]  
\centering  
\caption{Accuracy and Macro-F1 of Fully Trained Methods on Eight Adversarial Attack Types (\%)}  
\label{tab:combined_results}  
\sisetup{  
  table-format=99.999999,   
  table-number-alignment=center 
}  
\begin{tabular}{l|cccccccc}  
\toprule  
Method & \multicolumn{8}{c}{Accuracy (\%)} \\   
\cmidrule{2-9}  
       & BIM & FGSM & PGD & RFGSM & DIFGSM & MIFGSM & TIFGSM & TIPIM \\   
\midrule  
Adapter        & 99.84 & 99.82 & 99.84 & 99.84 & 99.84 & 99.84 & 99.84 & 99.77 \\   
Prompt Tuning  & 99.51 & 99.49 & 99.51 & 99.51 & 99.51 & 99.51 & 99.49 & 99.44 \\   
Fusion Model   & 99.87 & 99.87 & 99.87 & 99.87 & 99.87 & 99.87 & 99.87 & 99.80 \\   
\midrule  
Method & \multicolumn{8}{c}{Macro-F1 (\%)} \\   
\cmidrule{2-9}  
       & BIM & FGSM & PGD & RFGSM & DIFGSM & MIFGSM & TIFGSM & TIPIM \\   
\midrule  
Adapter        & 99.37 & 99.29 & 99.37 & 99.37 & 99.37 & 99.37 & 99.37 & 99.11 \\   
Prompt Tuning  & 98.08 & 97.98 & 98.06 & 98.08 & 98.06 & 98.08 & 98.00 & 97.79 \\   
Fusion Model   & 99.51 & 99.51 & 99.51 & 99.51 & 99.51 & 99.51 & 99.51 & 99.22 \\   
\bottomrule  
\end{tabular}  
\end{table}  

\begin{table}[htbp]
\renewcommand{\arraystretch}{1.0}
  \centering
  \caption{The cross-attack generalization performance of the proposed method. Each row represents the model trained on one specific adversarial example type and tested on all eight adversarial example types. }
  \resizebox{\textwidth}{!}{
        \setlength{\tabcolsep}{1.5mm}{
    \begin{tabular}{c|cccccccc}
    \hline
          & \multicolumn{8}{c}{Test set Accuracy (\%)} \\
\cline{2-9}    Training set & BIM   & FGSM  & PGD   & RFGSM & DIFGSM & MIFGSM & TIFGSM & TIPIM \\
    \hline
    BIM   & 99.81 & 99.74 & 99.74 & 99.80  & 99.60  & 99.84 & 99.17 & 96.47 \\
    FGSM  & 99.85 & 99.95 & 99.80  & 99.81 & 99.53 & 99.94 & 98.39 & 94.47 \\
    PGD   & 99.88 & 99.84 & 99.82 & 99.85 & 99.70  & 99.89 & 99.23 & 96.51 \\
    RFGSM & 99.81 & 99.74 & 99.73 & 99.80  & 99.62 & 99.84 & 99.09 & 96.17 \\
    DIFGSM & 99.87 & 99.73 & 99.77 & 99.82 & 99.72 & 99.82 & 99.58 & 98.28 \\
    MIFGSM & 99.81 & 99.77 & 99.72 & 99.78 & 99.53 & 99.87 & 98.72 & 95.19 \\
    TIFGSM & 98.41 & 98.45 & 98.33 & 98.37 & 98.28 & 98.53 & 99.78 & 99.09 \\
    TIPIM & 97.76 & 98.84 & 98.13 & 98.00  & 97.58 & 98.16 & 99.34 & 99.50 \\
    \hline
          & \multicolumn{8}{c}{Test set Macro-F1 (\%)} \\
\cline{2-9}    Training set & BIM   & FGSM  & PGD   & RFGSM & DIFGSM & MIFGSM & TIFGSM & TIPIM \\
    \hline
    BIM   & 99.28 & 99.02 & 99.02 & 99.26 & 98.53 & 99.41 & 96.99 & 89.02 \\
    FGSM  & 99.44 & 99.81 & 99.23 & 99.28 & 98.28 & 99.76 & 94.44 & 84.39 \\
    PGD   & 99.55 & 99.39 & 99.31 & 99.44 & 98.86 & 99.57 & 97.20  & 89.12 \\
    RFGSM & 99.28 & 99.02 & 98.99 & 99.26 & 98.58 & 99.39 & 96.72 & 88.26 \\
    DIFGSM & 99.49 & 98.99 & 99.12 & 99.31 & 98.94 & 99.33 & 98.45 & 94.11 \\
    MIFGSM & 99.28 & 99.13 & 98.94 & 99.18 & 98.25 & 99.49 & 95.48 & 85.96 \\
    TIFGSM & 94.46 & 94.58 & 94.2  & 94.33 & 94.06 & 94.84 & 99.17 & 96.69 \\
    TIPIM & 92.42 & 95.80  & 93.53 & 93.14 & 91.89 & 93.62 & 97.53 & 98.12 \\
    \hline
    \end{tabular}%
    }}
  \label{tab:generalization}%
\end{table}%

\subsection{Generalization Performance}
To validate the generalization ability of our method to unknown attacks, we conduct cross-validation experiments across eight attack types (BIM, FGSM, PGD, RFGSM, DIFGSM, MIFGSM, TIFGSM, TIPIM). For each attack type used as the training set, the remaining attacks serve as test sets. Metrics include classification accuracy (Accuracy) and macro-F1 score.
The model demonstrates strong generalization after training on a single attack type. For example, when trained on FGSM, it achieves an average accuracy of 99.80\% and macro-F1 of 99.25\% across other gradient-based attacks (e.g., BIM, PGD), indicating its ability to capture generic adversarial patterns rather than overfitting local attack features. When trained on TIFGSM, it detects TIPIM with 99.50\% accuracy and 98.12\% macro-F1, verifying its capacity to handle complex perturbations. However, limitations emerge in extreme scenarios. For instance, training on BIM or FGSM yields macro-F1 scores of 89.02\% and 84.39\% for TIPIM, respectively, highlighting the distinctiveness of physical perturbations compared to digital attacks. In addition, models trained on DIFGSM/MIFGSM exhibit lower TIPIM detection performance (macro-F1 $\approx$ 94\%) due to disrupted local feature consistency, suggesting the need for enhanced feature robustness via adversarial training \cite{A3}.

\subsection{Training Efficiency}
\subsubsection{Parameter Analysis}
As a classical benchmark model in computer vision, ResNet-18 demonstrates significant advantages in feature extraction owing to its deep residual architecture. However, its full-parameter training mechanism (11.18M trainable parameters) inevitably introduces substantial computational redundancy. In stark contrast, the proposed multimodal fusion framework achieves a nearly 20-fold reduction in parameter count through dynamic parameter sharing and lightweight adaptation design, requiring only 0.59M trainable parameters. Notably, this improvement in parameter efficiency does not compromise performance degradation. In Table \ref{tab:model_comparison}, the fusion model exhibits superior classification accuracy over ResNet-18 under adversarial testing while incurring only a 34.8\% increase in training time. This synergistic optimization of high parameter efficiency and enhanced robustness validates the innovative potential of the dynamic fusion mechanism in balancing model complexity with task adaptability.

\begin{table}[htbp]
\centering
\caption{Comparison of Model Parameters, Training Time and Performance}
\label{tab:model_comparison}
\resizebox{\textwidth}{!}{
        \setlength{\tabcolsep}{3mm}{
\begin{tabular}{l c c c}

\toprule
\textbf{Model} & \textbf{Params (M)} & \textbf{Training Epochs} & \textbf{Accuracy (\%)} \\
\midrule
ResNet-18       & 11.18              & 1000                        & 97.92                  \\
Adapter         & 0.59               & 572                        & 99.82                  \\
Prompt Tuning   & 0.012              & 270                        & 99.49                  \\
Fusion Model    & 0.59               & 240                        & 99.87                  \\
\bottomrule
\end{tabular}}}

\begin{tablenotes}
\small
\item \textbf{Note}: All models are tested on 1999 clean and 8K adversarial face images under FGSM attack. The trainable ratio of parameters for ResNet-18 is 1.00\%, for Fusion Model is 0.15\%, for Adapter is 0.14\%, and for Prompt Tuning is 0.003\%.
\end{tablenotes}
\end{table}

\subsubsection{Training speed and stability}
\begin{figure}[htbp]
    \centering
    \includegraphics[width=1\textwidth]{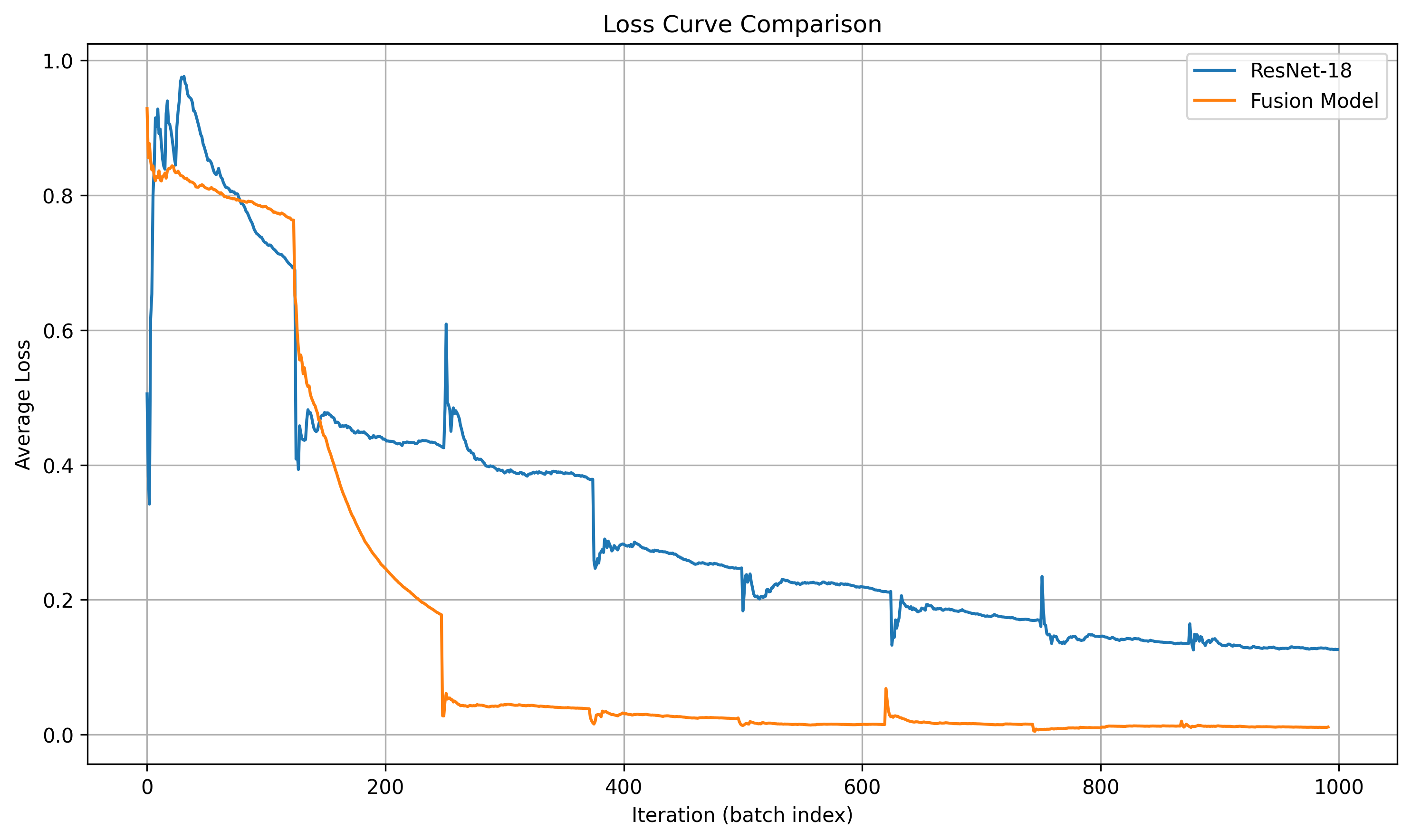}
    \caption{Comparison of training loss curves between a ResNet18 detector and ours.}
    \label{fig:train_loss}
\end{figure}

As illustrated in Figure \ref{fig:train_loss}, the comparative analysis of training dynamics demonstrates that the proposed fusion model exhibits significant advantages in both convergence efficiency and optimization stability. Through systematic examination of loss trajectories (Figure \ref{fig:train_loss}), the fusion model reveals a rapidly descending characteristic in its learning curve, achieving convergence more than twice as fast as the baseline ResNet-18 architecture. This accelerated convergence embodies the model's sophisticated parameter interaction mechanism, enabling effective gradient propagation through multi-modal feature pathways. Notably, the fused architecture attains a stable optimization plateau within 300 batches, 50\% earlier than its ResNet-18 counterpart, while maintaining superior loss suppression precision. This dual advantage of rapid stabilization and deep minimization originates from its adaptive feature recalibration module, which dynamically adjusts learning weights to prevent oscillatory traps that commonly plague conventional architectures.

\begin{table}[htbp]
\renewcommand{\arraystretch}{1.0}
  \centering
  \caption{The robustness performance of the proposed method to various perturbation magnitudes. Each row represents the model trained on one specific adversarial example type and tested on all eight adversarial example types. }
  \resizebox{\textwidth}{!}{
        \setlength{\tabcolsep}{1.5mm}{
    \begin{tabular}{c|cccccccc}
    \hline
          & \multicolumn{8}{c}{Test set Accuracy (\%)} \\
\cline{2-9}    Training set & BIM   & FGSM  & PGD   & RFGSM & DIFGSM & MIFGSM & TIFGSM & TIPIM \\
    \hline
    BIM    & 99.90 & 99.92 & 99.67 & 99.87 & 99.65 & 99.98 & 98.88 & 99.72 \\
    FGSM   & 99.92 & 99.99 & 99.89 & 99.91 & 99.60 & 99.99 & 98.21 & 99.67 \\
    PGD    & 99.90 & 99.96 & 99.76 & 99.87 & 99.68 & 99.98 & 98.91 & 99.67 \\
    RFGSM  & 99.89 & 99.92 & 99.64 & 99.84 & 99.61 & 99.98 & 98.70 & 99.71 \\
    DIFGSM & 99.88 & 99.91 & 99.72 & 99.87 & 99.70 & 99.96 & 99.51 & 99.84 \\
    MIFGSM & 99.87 & 99.95 & 99.62 & 99.82 & 99.56 & 99.97 & 98.44 & 99.66 \\
    TIFGSM & 99.00 & 99.86 & 99.41 & 99.00 & 98.44 & 99.84 & 99.87 & 99.91 \\
    TIPIM  & 98.85 & 99.82 & 99.70 & 98.97 & 98.30 & 99.76 & 99.65 & 99.85 \\
    \hline
          & \multicolumn{8}{c}{Test set Macro-F1 (\%)} \\
\cline{2-9}    Training set & BIM   & FGSM  & PGD   & RFGSM & DIFGSM & MIFGSM & TIFGSM & TIPIM \\
    \hline
    BIM    & 99.63 & 99.71 & 98.76 & 99.49 & 98.71 & 99.92 & 96.01 & 98.94 \\
    FGSM   & 99.68 & 99.95 & 99.60 & 99.65 & 98.53 & 99.95 & 93.87 & 98.76 \\
    PGD    & 99.63 & 99.84 & 99.10 & 99.49 & 98.79 & 99.92 & 96.10 & 98.76 \\
    RFGSM  & 99.60 & 99.71 & 98.66 & 99.41 & 98.56 & 99.92 & 95.44 & 98.92 \\
    DIFGSM & 99.55 & 99.65 & 98.94 & 99.49 & 98.86 & 99.87 & 98.20 & 99.39 \\
    MIFGSM & 99.52 & 99.81 & 98.58 & 99.31 & 98.38 & 99.89 & 94.60 & 98.74 \\
    TIFGSM & 96.39 & 99.46 & 97.83 & 96.39 & 94.57 & 99.41 & 99.49 & 99.65 \\
    TIPIM  & 95.86 & 99.32 & 98.85 & 96.26 & 94.07 & 99.08 & 98.69 & 99.43 \\
    \hline
    \end{tabular}%
    }}
  \label{tab:robustness}%
\end{table}%

\subsection{Robustness to Various Perturbation Magnitudes}
To evaluate the model's robustness against varying adversarial perturbation intensities ($\epsilon$ values), we conduct a comparative analysis between weak ($\epsilon=5/255$) and strong ($\epsilon=10/255$) perturbation conditions. Compare Table \ref{tab:generalization} with Table \ref{tab:robustness}, 
The experimental protocol employs cross-validation with one attack type allocated for training and remaining types for testing, using classification accuracy and Macro-F1 score as evaluation metrics. Results demonstrate that our fused architecture maintains high recognition accuracy across different perturbation intensities while preserving superior generalization capability. Specifically, the model achieves less than 2\% performance degradation when $\epsilon$ increases from 5/255 to 10/255, indicating robust feature disentanglement against adversarial distortions.

\section{CONCLUSION}
This study investigates CLIP’s robustness in detecting adversarially manipulated facial images. Using 2k clean faces from CelebA and generating 8k adversarial counterparts, we explore two transfer learning strategies: prompt tuning, adapter networks, and a fusion architecture. Comprehensive evaluation across eight attack types demonstrates the fusion architecture’s superior generalization. Our results highlight the advantages of multimodal transfer learning (combining visual and textual features) over conventional baselines, achieving significant improvements while minimizing training time. Cross-attack generalization experiments and robustness analyses further confirm its effectiveness under diverse perturbations. Future work will focus on refining prompt tuning to reduce parameter costs while maintaining performance.

%
%

\bibliographystyle{splncs04}
\bibliography{References}

\end{document}